\documentclass[11pt,a4paper]{article}
\usepackage[utf8]{inputenc}
\usepackage[T1]{fontenc}
\usepackage{amsmath,amssymb}
\usepackage{graphicx}
\usepackage{booktabs}
\usepackage{hyperref}
\usepackage{xcolor}
\usepackage{geometry}
\usepackage{natbib}
\usepackage{multirow}
\usepackage{caption}
\usepackage{placeins}

\title{FinSheet-Bench: From Simple Lookups to Complex Reasoning, Where LLMs Break on Financial Spreadsheets}

\author{
Jan Ravnik$^{1}$\thanks{Corresponding author: \texttt{info@qubera.ch}}, Matja\v{z} Li\v{c}en$^{1}$, Felix B\"uhrmann$^{1}$, Bithiah Yuan$^{2}$, \\
Felix Stinson$^{1}$, and Tanvi Singh$^{1}$ \\
\\
$^{1}$Qubera AG, Gartenstrasse 33, 8002 Z\"urich \\
$^{2}$University of Zurich, Department of Informatics, Binzm\"uhlestrasse 14, 8050 Z\"urich \\
}

\date{March 2026}

\begin{document}

\maketitle

\begin{abstract}
While Large Language Models (LLMs) can accelerate text-heavy tasks in alternative investment due diligence, a gap remains in their ability to accurately extract and reason over structured tabular data from complex financial spreadsheets. Progress is held back by the lack of real industry fund portfolio datasets for benchmarking, as private equity data rooms are confidential. To address this, we introduce \textbf{FinSheet-Bench}, a benchmark of synthetic financial portfolio data modeled on real private equity fund structures, designed to evaluate LLM performance on text-serialized spreadsheet question answering and numeric reasoning tasks.

Our evaluation of ten model configurations from OpenAI, Google, and Anthropic on financial spreadsheets, including complex layouts, fund dividers, and multi-line column names, reveals that \textbf{no standalone model achieves error rates low enough for unsupervised use} in professional finance applications. The best-performing model, Gemini 3.1 Pro, achieves 82.4\% accuracy across twenty-four evaluation files of varying complexity and structural layout (approximately 1~error per 6~questions), followed by GPT-5.2 with reasoning at 80.4\%, Claude Opus 4.6 with thinking at 80.2\%, and Gemini 3 Pro at 80.2\%. Performance degrades substantially on larger, more complex spreadsheets: the largest spreadsheet (152 companies, 8 funds) yields an average accuracy of just 48.6\% across all models, compared to 86.2\% on the easiest evaluation file. These difficulty patterns are consistent across all ten models, indicating that they reflect LLM limitations rather than idiosyncratic model weaknesses. Reliable financial spreadsheet extraction will likely require architectural approaches that separate document understanding from deterministic computation.
\end{abstract}

\section{Introduction}

Large Language Models (LLMs) such as GPT-5 \citep{openai2025gpt5}, Gemini 3.1 Pro \citep{google2026gemini31}, and Claude Opus 4.6 \citep{anthropic2026claude} now routinely match or exceed human performance on many text-based tasks. In the financial services industry, practitioners already use LLMs in document-heavy workflows: analyzing legal agreements, extracting key clauses from private placement memoranda, summarizing due diligence questionnaires, and drafting investment memos \citep{wu2023bloomberggpt, lee2025finllms, lopezlira2025bridging}.

\subsection{The Due Diligence Bottleneck}

Alternative investment due diligence remains one of the most resource-intensive processes in finance. Evaluating a single private equity or venture capital fund can require significant analyst effort and cost.\footnote{Industry practitioners report figures ranging from 55--120 analyst hours and \$20k--\$100k per fund evaluation, with total pre-investment costs of \$200k--\$1M per allocation. These estimates are based on practitioner experience rather than published surveys.} With global private markets assets under management now exceeding \$24 trillion \citep{mckinsey2026gpmr}, Limited Partners (LPs) need tools to scale diligence without sacrificing rigor.

While LLMs can significantly accelerate text-based diligence tasks (document review, clause comparison, compliance checking, and memo generation), a persistent challenge is the extraction of structured financial data from Excel spreadsheets. Portfolio monitoring spreadsheets, cash flow statements, and track record files are central to fund evaluation, yet these files have no standard format. They vary in sheet structure, column naming, header placement, data layout, and the use of formatting conventions that convey semantic meaning.

\subsection{The Spreadsheet Challenge}

Financial spreadsheets present several challenges that distinguish them from both unstructured text and simplified tabular benchmarks:

\begin{enumerate}
    \item \textbf{Complex multi-sheet structures}: Real financial models span multiple worksheets with complex cross-references, fund dividers, and section headers embedded within data rows.
    \item \textbf{Non-standard layouts}: Unlike clean CSV datasets, financial spreadsheets contain merged cells, hierarchical multi-row headers, embedded formulas, footnotes, and visual formatting that conveys semantic meaning.
    \item \textbf{Domain-specific conventions}: Financial terminology, calculation methodologies, and presentation standards vary across institutions and asset classes.
    \item \textbf{Low tolerance for errors}: In professional finance, even small numerical errors can have significant consequences. A 5\% error in a valuation metric could misrepresent millions in investment decisions. Industry practitioners typically expect very high accuracy (often cited as $\sim$97\%+) for automated financial workflows.\footnote{This threshold is an operational heuristic based on practitioner experience, not a formally defined regulatory standard. At 97\% accuracy on a 100-question extraction, approximately 3~answers would still be wrong, requiring human review. At the 82.4\% accuracy of our best model, roughly 1~in~6 questions is wrong, meaning the review burden may be approaching that of manual extraction. See Section~\ref{sec:limitations} for further discussion.}
\end{enumerate}

\subsection{The Dataset Gap}

A key obstacle to progress in financial spreadsheet extraction is the lack of publicly available, real-world benchmark datasets. Private equity fund data rooms are confidential by nature. LPs sign non-disclosure agreements, and fund managers guard their portfolio data closely. This confidentiality makes it impossible to create public benchmarks from real data, leaving researchers without a standardized way to evaluate extraction systems.

Existing tabular benchmarks such as WikiTableQuestions \citep{pasupat2015compositional}, Spider \citep{yu2018spider}, and TabFact \citep{chen2020tabfact} use simplified, well-structured tables that do not capture the complexity of real financial spreadsheets. Even financial-specific benchmarks like FinQA \citep{chen2021finqa} and TAT-QA \citep{zhu2021tatqa} focus on public company filings rather than the messy, multi-sheet workbooks used in private markets.

\section{Related Work}

\subsection{LLMs on Tabular Data}

Prior work has explored LLM capabilities on tabular data through benchmarks like WikiTableQuestions \citep{pasupat2015compositional}, Spider \citep{yu2018spider}, and TabFact \citep{chen2020tabfact}, focusing on semantic parsing, SQL generation, and fact verification over relatively clean tabular data. More recent work has examined LLM performance on spreadsheet tasks specifically: SheetCopilot \citep{li2023sheetcopilot} and SpreadsheetLLM \citep{dong2024spreadsheetllm} propose specialized architectures for spreadsheet understanding, TableBench \citep{wu2025tablebench} reveals that even frontier LLMs fall short of human performance across 18~fields of table QA, and the SemEval-2025 shared task \citep{grijalba2025semeval} provides a recent community-wide evaluation of LLM tabular reasoning. Closest to our setting, TableLLM \citep{zhang2025tablellm} evaluates tabular data manipulation in real office scenarios, though on well-structured tables rather than the complex, multi-fund financial layouts we target.

These benchmarks share a common limitation: they use simplified, well-structured data that does not capture the complexity of real-world financial models. We focus specifically on private equity portfolio analysis, where spreadsheet structures are irregular and error tolerance is minimal.

\subsection{Financial NLP and Document Understanding}

Financial NLP spans sentiment analysis \citep{araci2019finbert}, document classification \citep{loukas2022finer}, and information extraction from financial reports \citep{chen2021finqa}. FinQA \citep{chen2021finqa} and TAT-QA \citep{zhu2021tatqa} specifically address question answering over financial documents with tables, but focus primarily on public company filings rather than the complex spreadsheet models used in private markets. Li et al.~\citep{li2025extracting} introduce a framework for automated financial data extraction from unstructured documents using LLMs, while Ke et al.~\citep{ke2025documentllm} survey the growing field of LLMs applied to document intelligence more broadly.

Recent work on financial document processing has highlighted the challenges of extracting structured data from unstructured or semi-structured sources \citep{wu2023bloomberggpt, lopezlira2025bridging}.

\subsection{Reasoning and Chain-of-Thought}

Chain-of-thought prompting \citep{wei2022chain} and its variants yield large gains on mathematical reasoning tasks; self-consistency \citep{wang2023selfconsistency} and tree-of-thought \citep{yao2023tree} extend these gains further.

\section{The FinSheet-Bench Dataset}

\subsection{Dataset Construction}

The synthetic data is derived from original portfolio monitoring spreadsheets sent by various General Partners (GPs) to family offices as part of the data room for fund investments. We distinguish two aspects of the construction process:

\begin{itemize}
    \item \textbf{Layout and templates}: The structural characteristics of each file (sheet organization, column names, header placement, fund dividers, formatting conventions) are modeled on the 8 real source spreadsheets. These structural templates preserve the realistic complexity that makes financial spreadsheet extraction challenging.
    \item \textbf{Cell values}: All cell values were regenerated through the transformation procedure described below. No original cell values appear in the dataset.
\end{itemize}

\noindent This means the dataset is best described as \emph{synthetic data generated from real structural templates}, rather than ``anonymized'' real data. The transformations (shifting, scaling, perturbation) are designed to prevent reconstruction of original values. The structural templates themselves could in principle reveal information about the source files' layouts, though not about the underlying financial data.

\subsubsection{Transformation of General Information and Formatting}

The following transformations were applied to remove all identifying information from each source file:

\begin{itemize}
    \item The name of the original fund was replaced by a generic one.
    \item Any notes, watermarks, and larger pieces of text were regenerated to appear similar to the original text but not contain any identifying information. When this was not possible, the text was removed.
    \item Names of persons in the spreadsheet were replaced with new fictitious ones.
\end{itemize}

\subsubsection{Row-Level Data Generation}

The data consists of individual rows, each containing information pertaining to a single investment. To keep the data structurally consistent with the original spreadsheet while ensuring it contains no real information, a new row was generated based on each original row using the following procedure:

\begin{enumerate}
    \item \textbf{Company identity}: New fictitious companies were generated with data including the company name, headquarters location, and a description matching the name. When required, the company was also assigned to an appropriate industrial sector matching the categories in the original data or using the GICS Industry Classification Standard.
    
    \item \textbf{Realized/unrealized status}: The status was copied from the original row, because depending on the status, specific cells are left blank. This pattern must be preserved in the generated row.
    
    \item \textbf{Dates}: New entry and exit dates were generated. If the status was unrealized, no exit date was given or the current date was used instead (depending on the original data formatting).
    
    \item \textbf{Categorical values}: Values in columns that are independent from the rest of the data (typically text inputs, e.g., whether a company is public or private) were randomly sampled from the values in the original data. When dependencies exist between categorical entries (e.g., ``Country'' and ``Local currency''), both values were copied from the same original row to maintain consistency, or one value was chosen at random and the other was matched manually.
    
    \item \textbf{Numerical values}: Two scaling factors $A$ and $B$ were randomly chosen from the interval $[0.5, 2.0]$. For each value, an additional perturbation factor $k$ was drawn uniformly from $[0.95, 1.05]$ to further obfuscate the relationship to the original data. Entry-related values were multiplied by $Ak$, exit-related values by $Bk$, and ratios between exit and entry values by $kB/A$. Values constrained to $[0, 1]$ (e.g., ownership stakes) were multiplied only by $k$ and capped at 1. All generated numerical values were rounded to match the precision of the original data.

    Note that this perturbation scheme does not explicitly preserve accounting identities (e.g., Assets $=$ Liabilities $+$ Equity, or Net Debt $=$ Total Debt $-$ Cash). Because entry and exit values are scaled by different factors ($A$ and $B$ respectively), and each value receives an independent perturbation $k$, relationships between values that should be arithmetically consistent in real data may be broken in the synthetic data. However, these relationships often do not hold in real data either, as they also contain minor inconsistencies in accounting identities. In cases where the values actually follow an exact equation, the same equation is used in the synthetic data as well to calculate the values exactly.
    
    \item \textbf{Other entities}: Names of people (e.g., board members) and other companies (e.g., co-investors) were replaced with new fictitious names.
\end{enumerate}

Once the original rows were anonymized using these rules, additional synthetic rows were generated using the same principles and appended to the data.

\subsubsection{Version Generation}

This procedure produced 8 Excel files, each containing more rows than the original file; these are denoted as ``version~A.'' From each version~A file, two additional variants (``version~B'' and ``version~C'') were derived, for a total of 24~files. In versions~B and~C, the underlying data was left mostly intact; however, approximately one-third of the rows were removed at random, and structural modifications were applied: empty columns were added or removed, data columns were removed, merged, or split (e.g., ``Industrial sector'' and ``Subsector'' separated into two columns), and list separators were changed (e.g., commas to semicolons). The specific modifications for each version are detailed in Appendix~\ref{app:version_details}.

\subsubsection{Ground Truth Generation}

For each synthetic spreadsheet, standardized tables were manually extracted containing only the relevant data fields in a consistent format, removing formatting artifacts such as merged cells and layout variations. Using these standardized tables, ground truth answers were computed via deterministic Python functions that extract values and perform calculations with full numeric precision, ensuring reproducible and verifiable expected values for all benchmark questions.

\section{Experimental Setup}

\subsection{Models Evaluated}

We evaluated ten model configurations spanning three providers. The complete list is shown in Table~\ref{tab:models}.

\begin{table}[htbp]
\centering
\caption{Models evaluated in this study, with approximate public release dates. No provider publicly discloses parameter counts; we therefore report only the model name, provider, and release date.}
\label{tab:models}
\begin{tabular}{lll}
\toprule
\textbf{Model} & \textbf{Provider} & \textbf{Release Date} \\
\midrule
GPT-3.5-Turbo & OpenAI & Jan 2024 \\
GPT-4o & OpenAI & May 2024 \\
GPT-5.2 (no reasoning) & OpenAI & Dec 2025 \\
GPT-5.2 (reasoning) & OpenAI & Dec 2025 \\
\midrule
Gemini 2.0 Flash & Google & Feb 2025 \\
Gemini 2.5 Pro & Google & Jun 2025 \\
Gemini 3 Pro & Google & Nov 2025 \\
Gemini 3.1 Pro & Google & Feb 2026 \\
\midrule
Claude Opus 4.6 & Anthropic & Feb 2026 \\
Claude Opus 4.6 (thinking) & Anthropic & Feb 2026 \\
\bottomrule
\end{tabular}
\end{table}

\subsection{Question Categories}
\label{sec:question_categories}

We designed questions spanning seven categories of increasing complexity across four complexity levels, shown in Table~\ref{tab:question_classification}. Category describes \emph{what} operation is performed; Complexity describes \emph{how many} cognitive steps are required. The complexity levels were assigned based on the number of cognitive operations and data access patterns required:

\begin{itemize}
    \item \textbf{Low}: Single-step value lookup or boolean classification (1 column, no calculation). Even ``low'' tasks like Q1 may require scanning the entire sheet for fund boundaries.
    \item \textbf{Medium}: Multi-step operations with simple logic, such as filtering, list extraction, or multi-value lookups (2--3 columns, simple grouping).
    \item \textbf{High}: Aggregation and calculation tasks (sum, average, max/min, sorting) across multiple rows and 3+ columns.
    \item \textbf{Very High}: Multi-step statistical operations (median, percentile) requiring computed intermediate values and group-level aggregation.
\end{itemize}

\begin{table}[htbp]
\centering
\caption{Question classification by type and complexity. Each of the 16~question templates is mapped to one of the seven categories defined above.}
\label{tab:question_classification}
\small
\begin{tabular}{clcc}
\toprule
\textbf{Q\#} & \textbf{Question (abbreviated)} & \textbf{Category} & \textbf{Complexity} \\
\midrule
1 & How many funds are there? & Simple Lookup & Low \\
2 & How many companies are in each fund? & Counting & Medium \\
3 & Which fund is the latest? & Simple Lookup & Low \\
4 & List all companies in the newest fund & List Extraction & Medium \\
5 & List all companies sorted by entry EBITDA & Sorting & High \\
6 & Highest entry EBITDA company per fund? & Aggregation & High \\
7 & Which funds have unrealized investments? & Filtering & Medium \\
8 & How many unrealized investments per fund? & Counting & Medium \\
9 & Is [company] realized or unrealized? & Simple Lookup & Low \\
10 & Entry enterprise value for [company]? & Simple Lookup & Low \\
11 & Total unrealized capital per fund? & Aggregation & High \\
12 & Average entry EV per fund? & Aggregation & High \\
13 & Most recent exit date? & Aggregation & Medium \\
14 & Highest entry debt/EBITDA ratio? & Aggregation & High \\
15 & Average net debt at acquisition for [fund]? & Aggregation & High \\
16 & Median net debt/EBITDA for all funds & Complex Aggregation & Very High \\
\bottomrule
\end{tabular}
\end{table}

\subsection{Evaluation Protocol}

For each question:
\begin{enumerate}
    \item The full spreadsheet data (in comma-separated text format; see Section~\ref{sec:serialization}) was provided as context
    \item A single question was posed to the model
    \item The model's response was evaluated using a cascading verification system (see Section~\ref{sec:verification} for full specification).
\end{enumerate}

\subsubsection{Cascading Verification System}
\label{sec:verification}

We employ a three-tier cascading verification system for the automated evaluation of free-text LLM answers against structured ground-truth values.

\paragraph{Tier~1: Exact matching (resolves $\sim$25\% of answers).}
Strict rule-based extraction is attempted first. A result is accepted at this tier only if confidence $\geq 0.95$.
\begin{itemize}
    \item \textbf{Numeric answers}: A number is extracted via strict regex (requiring clear word boundaries) and compared with 2.5\% relative tolerance (half of the 5\% tolerance used in Tier~2). We chose 5\% as the primary tolerance because ground-truth values in our dataset are rounded to 1--2 decimal places (e.g., multiples to 1 decimal, currency values to nearest million), and models frequently reproduce values at different precision (e.g., ``12.3'' vs.\ ``12.34'').
    \item \textbf{String answers}: Case-insensitive exact match after whitespace normalization (e.g., ``Fund III'' matches ``FundIII'').
    \item \textbf{List answers}: Items are extracted, lowercased, and compared as sets using Jaccard similarity (intersection over union) with a strict threshold of 0.95. Comparison is order-insensitive for all list questions.
    \item \textbf{Dictionary answers} (e.g., ``Fund~I: 5, Fund~II: 3''): Keys are matched and each value is verified independently; the answer is accepted if $\geq$95\% of key-value pairs match.
    \item \textbf{Date answers}: Parsed and compared with 1-day tolerance to handle timezone and time-of-day formatting differences.
    \item \textbf{Boolean answers}: Keyword detection (e.g., ``realized''/``unrealized'', ``yes''/``no'').
\end{itemize}

\paragraph{Tier~2: Fuzzy matching (resolves $\sim$25\%).}
When Tier~1 confidence is below 0.95, a more tolerant rule-based pass is applied. A result is accepted at this tier if confidence $\geq 0.70$.
\begin{itemize}
    \item \textbf{Numeric answers}: Broader regex extraction (any number in text) with the full 5\% relative tolerance, i.e., $|a - e| / |e| < 0.05$.
    \item \textbf{String answers}: Partial/substring matching (ground truth contained in answer) and sequence similarity via Python's \texttt{difflib.SequenceMatcher} (Ratcliff/Obershelp algorithm) with a threshold of 0.95.
    \item \textbf{List answers}: Jaccard set similarity with a relaxed threshold of 0.75. Items are compared as sets (order-insensitive; duplicates are not preserved).
    \item \textbf{Dictionary answers}: Key-value pair matching with a relaxed threshold of 0.75.
\end{itemize}

\paragraph{Tier~3: LLM adjudication (resolves $\sim$48\%).}
LLM adjudication (GPT-4o-mini and Gemini~3~Flash~Preview) is invoked when Tiers~1 and~2 produce confidence below 0.70, which occurs frequently for multi-field and list-type answers where free-text responses are difficult to parse with rules alone. The adjudicator receives both the expected and actual answers with a structured Pydantic schema requiring a boolean verdict, confidence score, extracted value, and explanation. The LLM adjudicator handles word-number conversion (e.g., ``four'' $\to$ 4), semantic equivalence of differently phrased answers, and complex formatting variations that rule-based tiers cannot resolve. All 2{,}332 LLM-adjudicated answers were independently checked using both models. The two models agreed on 93.1\% of answers (2{,}171/2{,}332); the 161 discrepant cases (6.9\%) were manually reviewed and resolved.

\begin{table}[htbp]
\centering
\caption{Verification tier distribution by answer type. Values are approximate percentages of answers resolved at each tier, aggregated across all model and question evaluations ($10 \times \sim\!500 \approx 5{,}000$ total). The high proportion of LLM adjudication reflects the difficulty of rule-based parsing of free-text responses, particularly for multi-field (dict) and list answers.}
\small
\begin{tabular}{lccc}
\toprule
\textbf{Answer Type} & \textbf{Tier~1 (Exact)} & \textbf{Tier~2 (Fuzzy)} & \textbf{Tier~3 (LLM)} \\
\midrule
Numeric/Dict (Q1--2, Q6, Q8, Q10--12, Q14--16) & 15\% & 30\% & 51\% \\
String/Date/Bool (Q3, Q9, Q13) & 60\% & 10\% & 26\% \\
List (Q4, Q5, Q7) & 4\% & 19\% & 73\% \\
\midrule
\textbf{Overall (all types)} & \textbf{$\sim$25\%} & \textbf{$\sim$25\%} & \textbf{$\sim$48\%} \\
\bottomrule
\end{tabular}
\begin{minipage}{\linewidth}
\vspace{2pt}
\footnotesize The remaining $\sim$3\% of evaluations (132 of 5{,}010) are GPT-3.5-Turbo context-overflow errors on files exceeding its 16K token limit; these produced no model answer and are excluded from the tier percentages.
\end{minipage}
\end{table}

\subsubsection{Question Sampling and Dataset Coverage}

The 16 question templates include three parameterized templates (Q9, Q10, Q15) that generate one question per entity (company or fund). Because running every entity for every model is prohibitively expensive, we sample a fixed number of entities per template (\texttt{sample\_size=3}, \texttt{seed=42}) to keep cost manageable while ensuring reproducibility. The total number of evaluated questions therefore varies depending on which data files were benchmarked:

All models were evaluated on 24 evaluation files (8 version~A, 8 version~B, and 8 version~C structural variants) with \texttt{sample\_size=3} and \texttt{seed=42}, yielding up to 22 questions per file (approximately 500 total per model). Per-file question counts vary because some templates require columns not present in every file, and version~B and~C files have fewer applicable templates due to removed columns (see Table~\ref{tab:file_results} for file sizes and per-file statistics). GPT-3.5-Turbo could not process 6 of 24 files due to its 16K context limit; those questions were treated as errors.

All comparisons in the paper are made on per-question accuracy (percentage correct) to normalize across different totals.

\subsection{Data Format}

All evaluation files are provided in a single text format: each Excel workbook is converted to a plain text file where sheets are separated by \texttt{=== Sheet: <name> ===} headings, with cell values rendered in comma-separated (CSV-like) format. This is the only serialization format used in all experiments reported in this paper. Converting the benchmark files to alternative representations (structured JSON, tabular HTML, or other formats) is a natural direction for future format-sensitivity studies (see Section~\ref{sec:limitations}).

\subsection{Prompt Design}

We deliberately used identical, minimal prompts for all models to ensure that performance differences reflect model capabilities rather than prompt optimization, to mirror real-world deployment where maintaining model-specific prompts is impractical, and to ensure reproducibility.

\subsubsection{Spreadsheet Serialization}
\label{sec:serialization}

A key methodological decision, and a central limitation of this study, is how raw Excel spreadsheets are converted to text for LLM consumption. We deliberately chose text/CSV serialization because it reflects how LLMs are overwhelmingly used in practice for spreadsheet extraction today. Text serialization is the de facto industry standard for feeding tabular data to language models: commercial RAG pipelines, document processing systems, and the research benchmarks cited in Section~2 all serialize tables as text. Evaluating this modality therefore has direct practical relevance, as it benchmarks the approach that practitioners actually deploy. Further, our results suggest that serialization-independent reasoning failures dominate the accuracy gap; see Section~6.2 for a detailed analysis of modality-dependent versus modality-independent failure modes.

\paragraph{Serialization Specification.}
Each \texttt{.xlsx} workbook is processed sheet-by-sheet using the \texttt{openpyxl} library in \texttt{data\_only=True} mode (formula cells return their cached value, not the formula). The steps are:

\begin{enumerate}
    \item \textbf{Cell reading}: Every cell is read left-to-right, top-to-bottom, using \texttt{openpyxl}'s \texttt{iter\_rows\allowbreak(values\_only\allowbreak=True)}. \texttt{None} cells are emitted as empty strings. All other values are cast to \texttt{str()}.
    \item \textbf{Merged cells}: \texttt{openpyxl} reports only the top-left value of a merged region; all other cells in the merge are \texttt{None} (rendered as empty). No special handling is applied.
    \item \textbf{Hidden rows/columns}: Not filtered; all rows and columns appear in the output regardless of Excel visibility settings.
    \item \textbf{Row numbers}: Not included. The model sees no row or column indices.
    \item \textbf{Number formatting}: Cell values are Python-native: floats remain as-is (e.g., \texttt{1200000.0}, not \texttt{\$1.2M}); dates appear as \texttt{datetime} objects cast to their string representation (e.g., \texttt{2024-06-15 00:00:00}).
    \item \textbf{Trailing cleanup}: Trailing all-empty columns and trailing all-empty rows are stripped per sheet.
    \item \textbf{Output format}: Each row is written with Python's \texttt{csv.writer} (comma-separated, RFC~4180 quoting). Multi-line cell values (e.g., ``Enterprise\textbackslash nValue'') are enclosed in double quotes with embedded newlines preserved.
    \item \textbf{Sheet boundaries}: All sheets in the workbook are placed in a single text file. Each sheet is preceded by a separator line: \texttt{=== Sheet: <name> ===}. Sheets are concatenated with a blank line between them.
\end{enumerate}

\noindent The model therefore sees a single text document containing all non-empty sheets in workbook order, with comma-separated columns, one row per line, and no row/column indices. Empty rows (e.g., fund dividers) are preserved as blank CSV lines.

\paragraph{Example.} A minimal $3 \times 4$ region serializes as:

\begin{verbatim}
=== Sheet: Portfolio Company Data Sheet ===
,,,,
,Nebula Portfolio Company Datasheet,,,,
,$ in millions,,,,
,,,,
,,,,
,Company,"Initial Investment
Date",Industry Vertical,Status
,Apex Inc,2022-01-15 00:00:00,Technology,Realized
,Bloom Ltd,2023-06-01 00:00:00,Healthcare,Unrealized
\end{verbatim}

\noindent Note that the header ``Initial Investment\textbackslash nDate'' spans two lines inside quotes, empty cells produce consecutive commas, and title/subtitle rows appear as data rows with most columns empty.

\paragraph{Information loss.}
This serialization discards several information channels present in the original Excel files: merged cells are flattened (only the top-left value survives), cell background colors and font formatting (bold, underline, italic) are lost, table borders and boundary lines are removed, cross-sheet references are not preserved, and spatial grouping (column widths, indentation) is collapsed. These visual cues are not merely aesthetic. In financial spreadsheets, bold text typically denotes subtotals or section headers, underlines separate data blocks, cell background colors distinguish fund sections, and table borders delineate data regions from annotations. Fund dividers and section headers are preserved as text rows but lose their visual distinctiveness, becoming indistinguishable from data rows. This information loss directly contributes to the failure modes (row misidentification, incorrect column selection). Vision-capable models processing rendered images of the spreadsheet could potentially recover these cues. Serialized file sizes vary substantially: the smallest variant B/C files contain approximately 7K characters ($\sim$5K tokens for GPT, $\sim$9K for Gemini), while the largest A-version file (synthetic4\_A, 152 companies) reaches approximately 108K characters ($\sim$44K GPT tokens, $\sim$63K Gemini tokens).

Because of this lossy serialization, the results in this paper reflect LLM performance on \emph{text-serialized spreadsheet QA}, not on native Excel comprehension. If models were given structured JSON, tabular HTML, or image-based representations, results might differ. Investigating alternative serialization formats is a direction for future work. All results reported here use the comma-separated text serialization described above. The ``spatial'' failures discussed in Section~6.2 are therefore partially attributable to the serialization method removing the 2D grid structure before the model processes the data.

\subsubsection{Prompt Template}

The prompt template used for all models was:

\begin{verbatim}
System: You are analyzing financial data for portfolio 
companies and funds.

DATA:
{spreadsheet_content}

Please provide direct, concise answers based on the 
data provided.

User: {question}
\end{verbatim}

No chain-of-thought prompting, few-shot examples, or other prompt engineering techniques were applied. This represents a ``zero-shot'' baseline evaluation. We do vary vendor-supported inference-time reasoning settings (OpenAI \texttt{reasoning\_effort}, Anthropic \texttt{extended\_thinking}), which are model configuration parameters rather than prompt modifications. Prompt optimization could potentially improve individual model scores, but would confound the comparison across models and reduce external validity.

\subsection{Metrics}

We evaluate each model on three dimensions: \textbf{accuracy} (percentage of correct answers), \textbf{response time} (average seconds per question), and \textbf{token usage} (input and output tokens consumed). Accuracy is the primary metric, but response time and token usage matter for practitioners because they directly determine API cost and throughput: a model that is 2\% more accurate but 10$\times$ slower or 5$\times$ more expensive per query may not be the right choice for processing hundreds of spreadsheets under time and budget constraints.

\section{Results}

\subsection{Overall Model Performance}

\begin{figure}[htbp]
\centering
\includegraphics[width=0.9\textwidth]{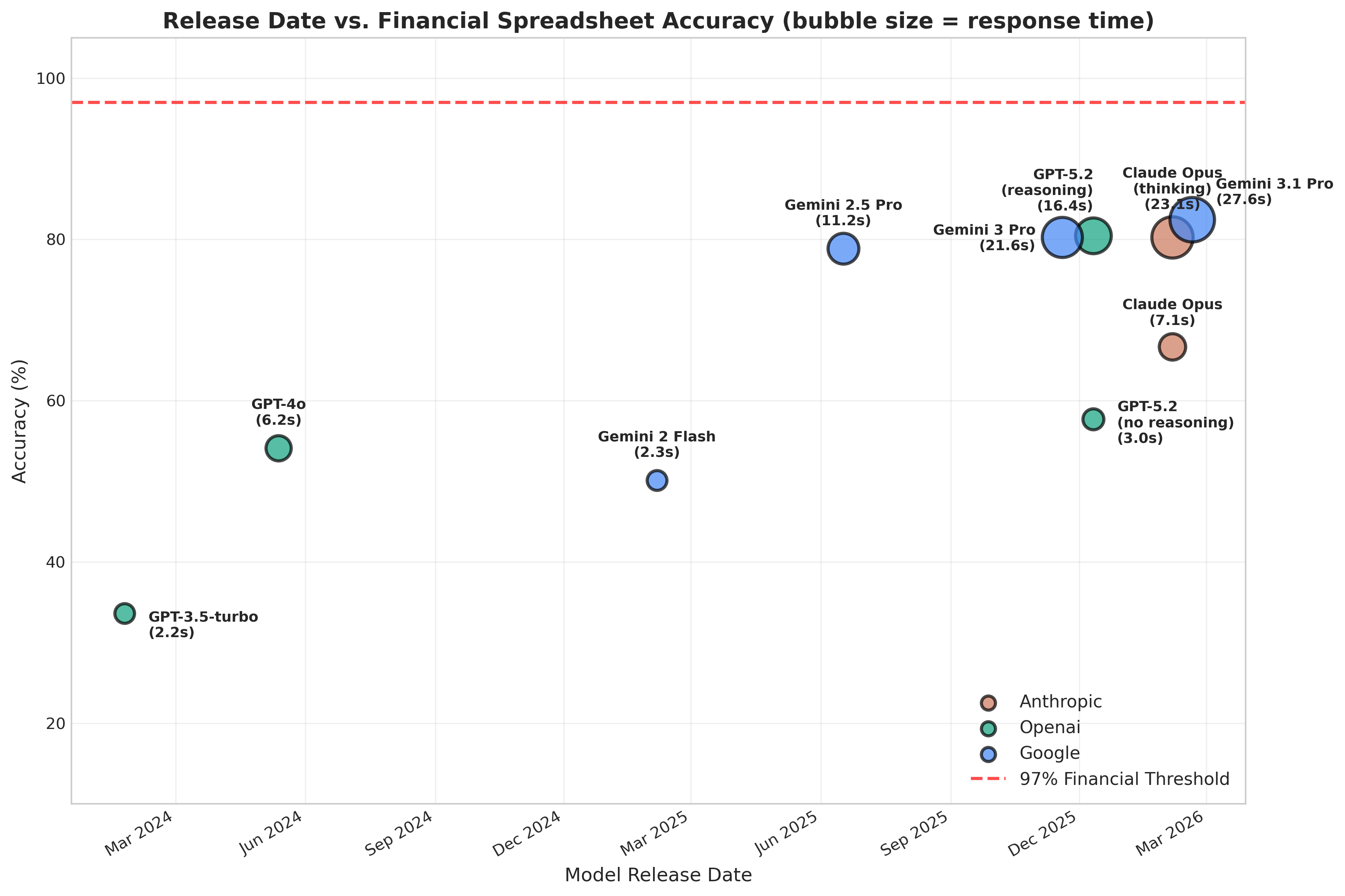}
\caption{LLM accuracy on financial spreadsheet tasks plotted against model release date. Bubble size indicates average response time per question. At 82.4\%, even the best model produces approximately 1~error per 6~questions. Newer models and models with reasoning generally achieve higher accuracy. See Table~\ref{tab:overall_results} for detailed results.}
\label{fig:overall_accuracy}
\end{figure}

Figure~\ref{fig:overall_accuracy} presents the overall accuracy of each model on the FinSheet-Bench dataset. Table~\ref{tab:overall_results} provides detailed metrics.

\begin{table}[htbp]
\centering
\caption{Overall model performance on FinSheet-Bench. All models except GPT-3.5-Turbo achieve 100\% coverage (fraction of questions the model could process). Accuracy (Processed) is the primary comparison metric: accuracy computed on the subset of questions each model could process. Effective Accuracy = Accuracy (Processed) $\times$ Coverage; this composite metric is reported only for GPT-3.5-Turbo to capture its coverage penalty and should not be compared directly against other models' accuracy scores. For all models with 100\% coverage, Effective Accuracy equals Accuracy (Processed).}
\label{tab:overall_results}
\small
\begin{tabular}{lccccc}
\toprule
\textbf{Model} & \textbf{Cov.} & \textbf{Acc.} & \textbf{Eff. Acc.} & \textbf{Time (s)} & \textbf{Total Tokens} \\
\midrule
\textbf{Gemini 3.1 Pro} & 100\% & \textbf{82.4\%} & \textbf{82.4\%} & 27.60 & 13,455,183 \\
GPT-5.2 (reasoning) & 100\% & 80.4\% & 80.4\% & 16.37 & 8,861,018 \\
Claude Opus 4.6 (thinking) & 100\% & 80.2\% & 80.2\% & 23.14 & 9,759,287 \\
Gemini 3 Pro & 100\% & 80.2\% & 80.2\% & 21.62 & 13,575,897 \\
Gemini 2.5 Pro & 100\% & 78.8\% & 78.8\% & 11.17 & 13,122,140 \\
Claude Opus 4.6 & 100\% & 66.7\% & 66.7\% & 7.11 & 9,145,661 \\
GPT-5.2 (no reasoning) & 100\% & 57.7\% & 57.7\% & 3.04 & 8,353,847 \\
GPT-4o & 100\% & 54.1\% & 54.1\% & 6.17 & 8,474,293 \\
Gemini 2.0 Flash & 100\% & 50.1\% & 50.1\% & 2.27 & 12,533,984 \\
GPT-3.5-Turbo$^*$ & 73.7\% & 33.6\% & 24.8\% & 1.63 & 2,900,242 \\
\bottomrule
\multicolumn{6}{l}{\small $^*$GPT-3.5-Turbo exceeded its 16K context limit on 6 of 24 files.} \\
\multicolumn{6}{l}{\small Coverage = 369/501 = 73.7\%. Accuracy on those 369 questions = 33.6\% (124/369).}
\end{tabular}
\end{table}

Across all ten configurations, \textbf{no standalone LLM achieves error rates low enough for unsupervised use} in production financial workflows. The best-performing model, Google's Gemini 3.1 Pro, achieves 82.4\% accuracy on twenty-four evaluation files (approximately 1~error per 6~questions), followed by GPT-5.2 with reasoning at 80.4\%, Claude Opus 4.6 with thinking at 80.2\%, and Gemini 3 Pro at 80.2\%. Performance degrades on larger, more complex spreadsheets: synthetic4\_A (152 companies, 8 funds) yields much lower accuracy for most models, with GPT-5.2 with reasoning and Claude Opus 4.6 with thinking both dropping to 55\% on that file, while Gemini 3.1 Pro achieves 82\% there. File complexity is clearly a key driver of accuracy, though the newest models hold up better on the most complex files.

GPT-3.5-Turbo's low effective accuracy (24.8\%) is partly a coverage story: it could not process 6 of 24 files due to its 16K context limit (failures are distributed across A, B, and C variants). On the 369~questions it could process, GPT-3.5-Turbo achieves 33.6\% accuracy, still the lowest among all models.

\subsection{Accuracy by Question Category}

\begin{figure}[htbp]
\centering
\includegraphics[width=0.95\textwidth]{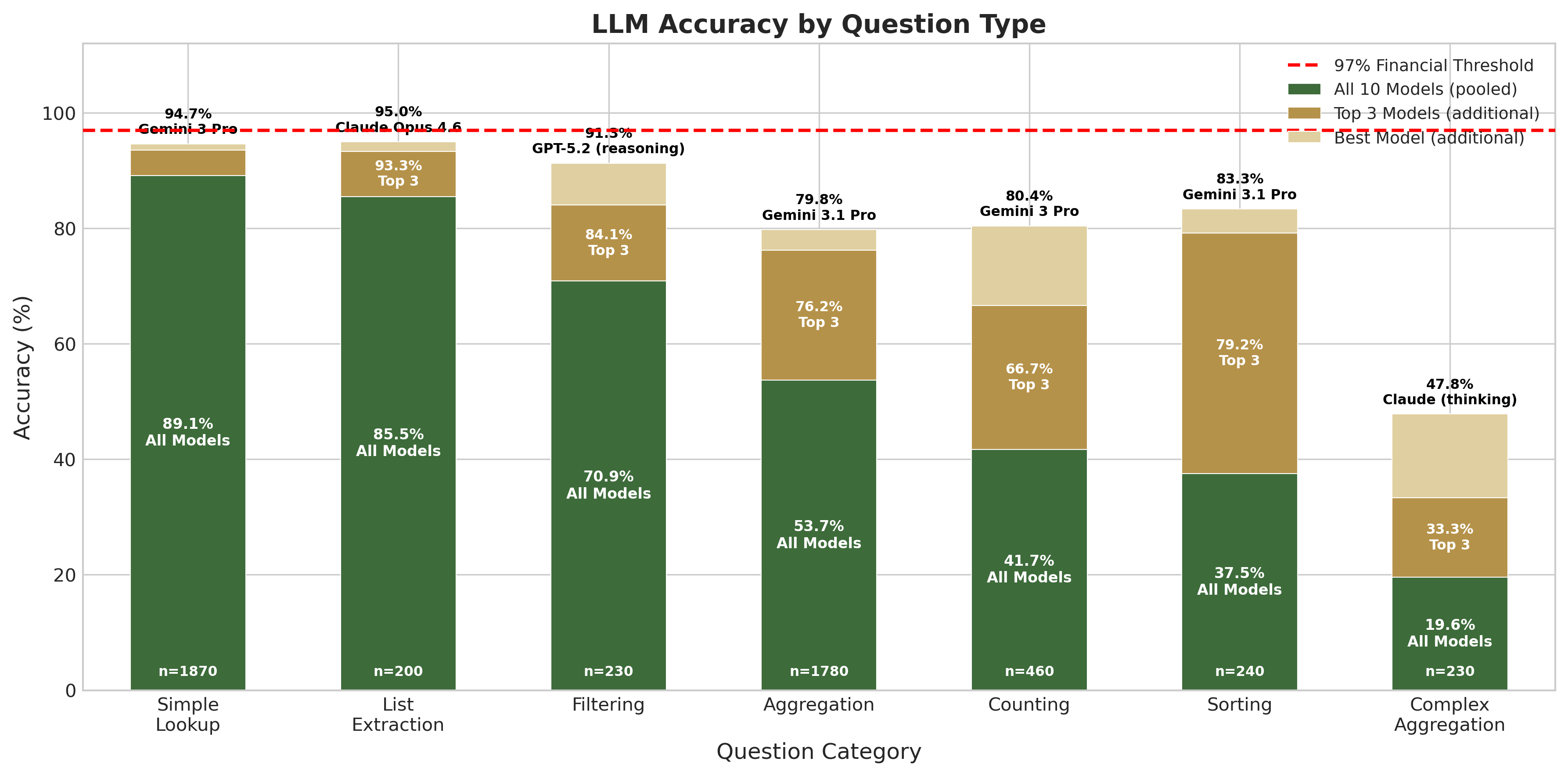}
\caption{LLM accuracy by question category (stacked bars). The dark base shows pooled accuracy across all ten models; the middle segment adds the gain when restricting to the top~3 models; the lightest segment adds the further gain of the single best-performing model per category (labeled above each bar). Simple lookups achieve high accuracy, but performance drops for tasks requiring calculation or multi-step reasoning. Sorting (37.5\% all models, 83\% best) is particularly challenging given that list extraction achieves 85.5\%; models can identify \emph{what} is in a list but struggle with \emph{ordering by magnitude}. See text for a detailed discussion of tokenization artifacts vs.\ reasoning failures.}
\label{fig:accuracy_by_type}
\end{figure}

Table~\ref{tab:question_types} and Figure~\ref{fig:accuracy_by_type} show accuracy broken down by the question categories. The results show a steep performance drop: while simple lookups generally achieve high accuracy (89.1\% overall, though individual models and questions vary), tasks requiring calculation and complex aggregation score much lower.

\begin{table}[htbp]
\centering
\caption{Pooled accuracy by question category. ``All Models'' is the pooled accuracy across all ten model configurations; ``Top~3'' is the pooled accuracy of the three best-performing models (Gemini~3.1 Pro, GPT-5.2 reasoning, Claude Opus~4.6 thinking).}
\label{tab:question_types}
\small
\begin{tabular}{llp{4.6cm}rrr}
\toprule
\textbf{Category} & \textbf{Complexity} & \textbf{Example} & \textbf{Attempts} & \textbf{All Models} & \textbf{Top 3} \\
\midrule
Simple Lookup & Low & How many funds are there? & 1,870 & 89.1\% & 93.6\% \\
List Extraction & Medium & List all companies in the newest fund & 200 & 85.5\% & 93.3\% \\
Filtering & Medium & Which funds have unrealized investments? & 230 & 70.9\% & 84.1\% \\
Aggregation & High & What is the average entry EV per fund? & 1,780 & 53.7\% & 76.2\% \\
Counting & Medium & How many unrealized investments per fund? & 460 & 41.7\% & 66.7\% \\
Sorting & High & List all companies sorted by entry EBITDA & 240 & 37.5\% & 79.2\% \\
Complex Agg. & Very High & Calculate median net debt/EBITDA for all funds & 230 & 19.6\% & 33.3\% \\
\bottomrule
\end{tabular}
\end{table}

This pattern is consistent across all models: tasks requiring multi-step numerical reasoning show much lower accuracy than simple information retrieval. Even ``simple'' tasks like Q1 (``How many funds are there?'') may require scanning the entire sheet to identify fund boundaries. Counting (Q2, Q8) is surprisingly difficult at 41.7\%, likely because it requires accurate fund boundary detection in serialized text.

\paragraph{Top-3 models only.}
Restricting the analysis to the three best-performing models (Gemini 3.1 Pro, GPT-5.2 with reasoning, Claude Opus 4.6 with thinking) provides a more practical view of current capabilities. These models achieve 93.6\% on simple lookups, 93.3\% on list extraction, 84.1\% on filtering, 76.2\% on aggregation, 66.7\% on counting, 79.2\% on sorting, and 33.3\% on complex aggregation. This difficulty gradient persists: even the best available models drop from $\sim$94\% on simple tasks to $\sim$33\% on complex aggregation. However, the absolute accuracy levels are substantially higher than the all-model averages, particularly for counting (66.7\% vs.\ 41.7\%) and sorting (79.2\% vs.\ 37.5\%), suggesting that the latest reasoning-enabled models have partially closed the gap on these challenging categories.

\begin{figure}[htbp]
\centering
\includegraphics[width=0.85\textwidth]{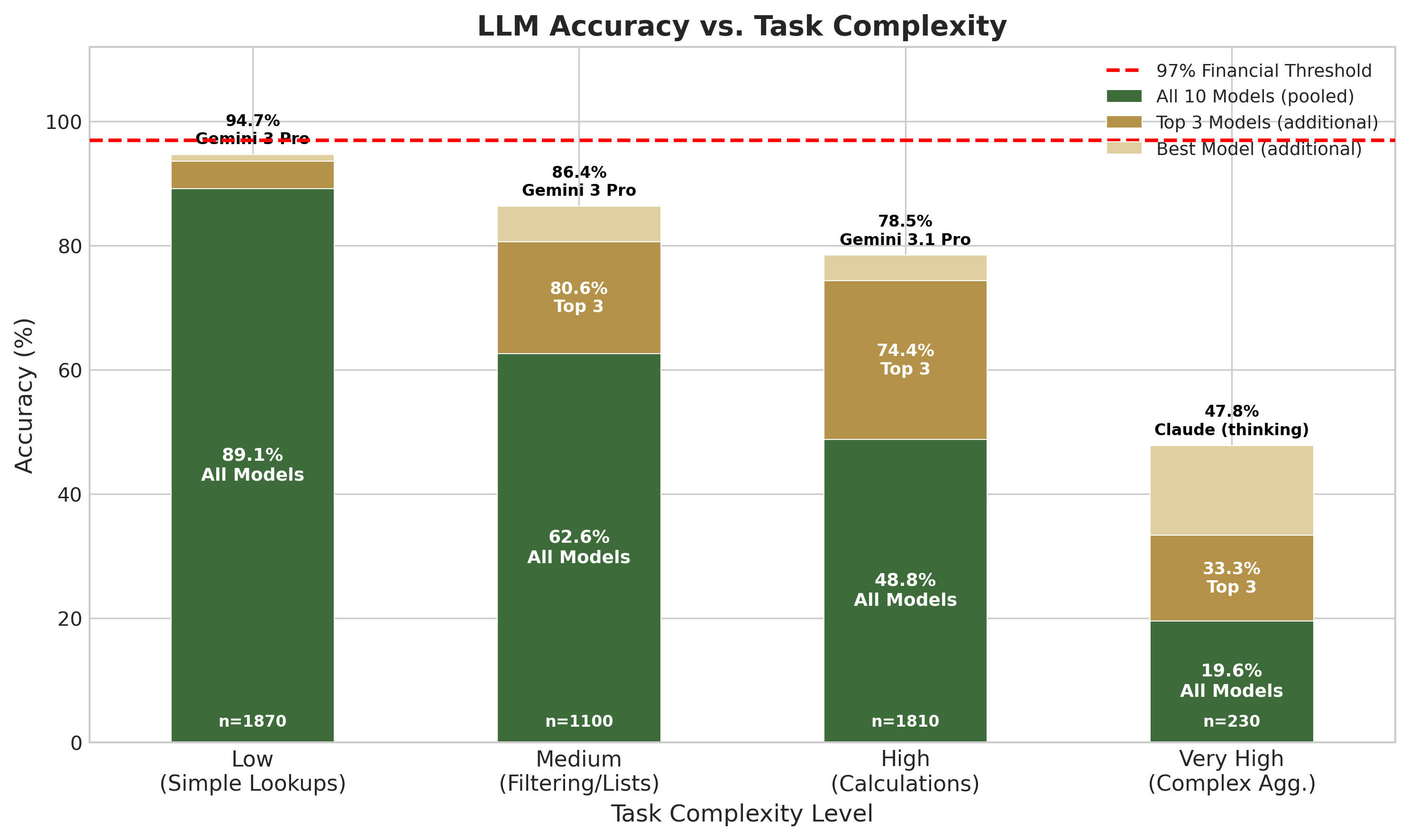}
\caption{Accuracy vs.\ task complexity level (stacked bars). The dark base shows pooled accuracy across all ten models; the middle segment adds the gain when restricting to the top~3 models; the lightest segment adds the further gain of the single best-performing model per complexity level (labeled above each bar). Low-complexity tasks achieve $\sim$89--95\% accuracy, but performance drops sharply at High and Very High complexity, where tasks require multi-step calculations, aggregations, or sorting. Even the best individual model drops to 47.8\% on Very High complexity.}
\label{fig:accuracy_by_complexity}
\end{figure}

Figure~\ref{fig:accuracy_by_complexity} visualizes the relationship between task complexity and accuracy. Accuracy drops steadily as task complexity increases, and the pattern holds across all ten models, pointing to a limitation of current LLM architectures rather than quirks of individual question categories.

\noindent For the top three models specifically, accuracy by complexity level is: Low~94\%, Medium~81\%, High~74\%, Very~High~33\%. The gradient is less steep than for all models combined (Low~89\% to Very~High~20\%), but the qualitative pattern is unchanged: even the best current models lose over 60~percentage points between simple and very complex tasks.

\begin{figure}[htbp]
\centering
\includegraphics[width=0.95\textwidth]{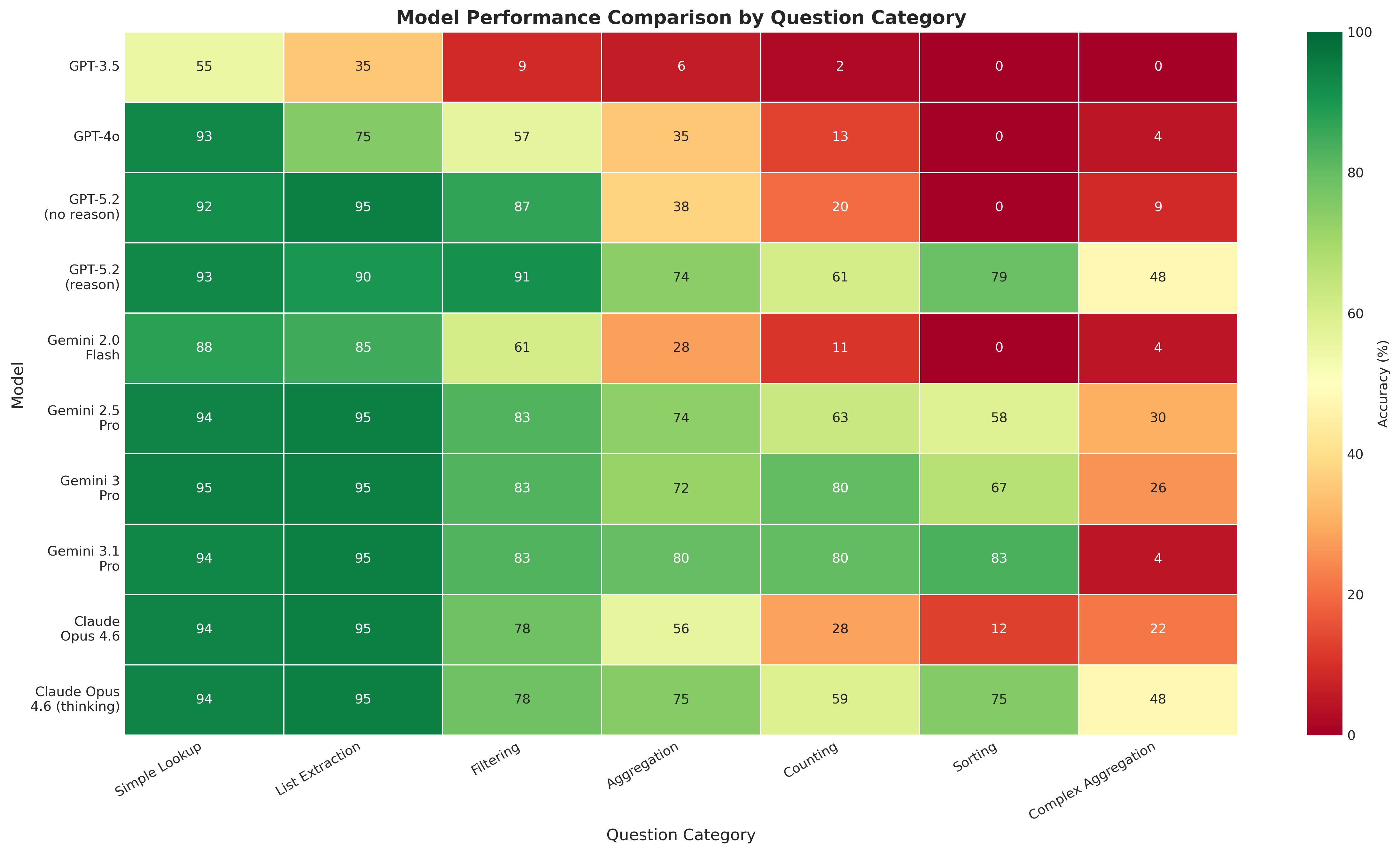}
\caption{Model performance comparison by question type across all ten model configurations (OpenAI, Google, Anthropic).}
\label{fig:model_comparison}
\end{figure}

Figure~\ref{fig:model_comparison} compares all ten model configurations across question categories. Earlier models struggle with most tasks; the latest reasoning-enabled models reach near-perfect accuracy on simple lookups but still fall short on complex aggregation and sorting, despite clear gains over their predecessors.

\begin{figure}[htbp]
\centering
\includegraphics[width=0.98\textwidth]{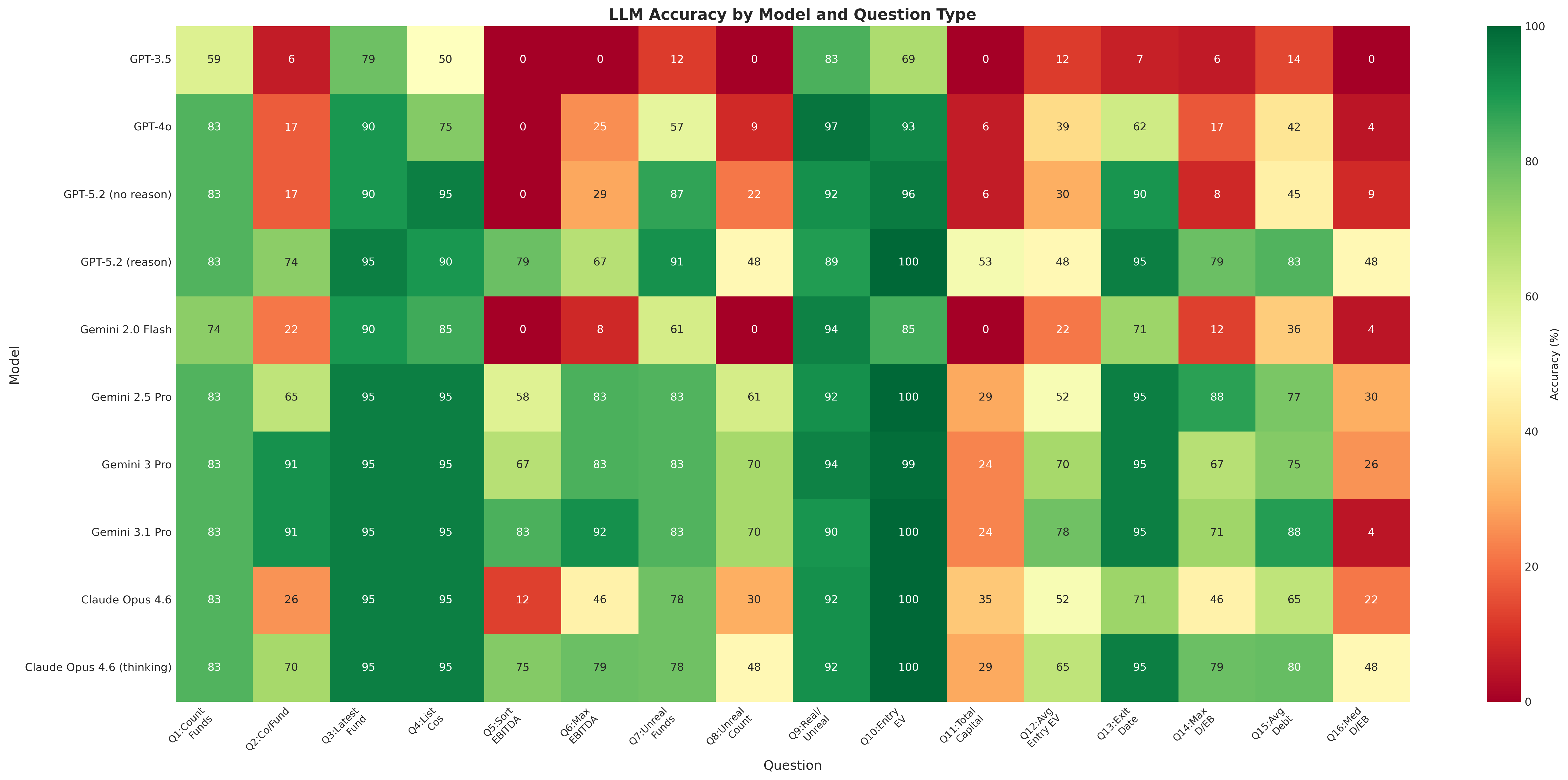}
\caption{Heatmap of accuracy by model and question type across all ten model configurations (4 OpenAI, 4 Google, 2 Anthropic). Green indicates high accuracy; red indicates failure. Results span twenty-four evaluation files (approximately 500 questions per model; some files contribute fewer questions due to missing columns; see Section~4.2 for per-file question counts).}
\label{fig:heatmap}
\end{figure}

Figure~\ref{fig:heatmap} provides a complete view of model performance across all question types, showing which tasks are hardest.

\subsection{The Reasoning Gap}

Enabling extended reasoning yields large accuracy gains. GPT-5.2 improves from 57.7\% to 80.4\% (+22.8pp) with reasoning enabled, while Claude Opus 4.6 improves from 66.7\% to 80.2\% (+13.6pp) with adaptive thinking.

Both models score markedly higher with reasoning turned on. However, vendors bundle multiple changes in ``reasoning mode'' (decoding constraints, verbosity constraints, internal scratchpad, tool use policies) that are not independently controllable. We therefore cannot isolate the contribution of chain-of-thought processing from these confounds.

Google's Gemini 3 Pro uses internal reasoning by default and we did not test its performance without reasoning.

\begin{figure}[htbp]
\centering
\includegraphics[width=0.95\textwidth]{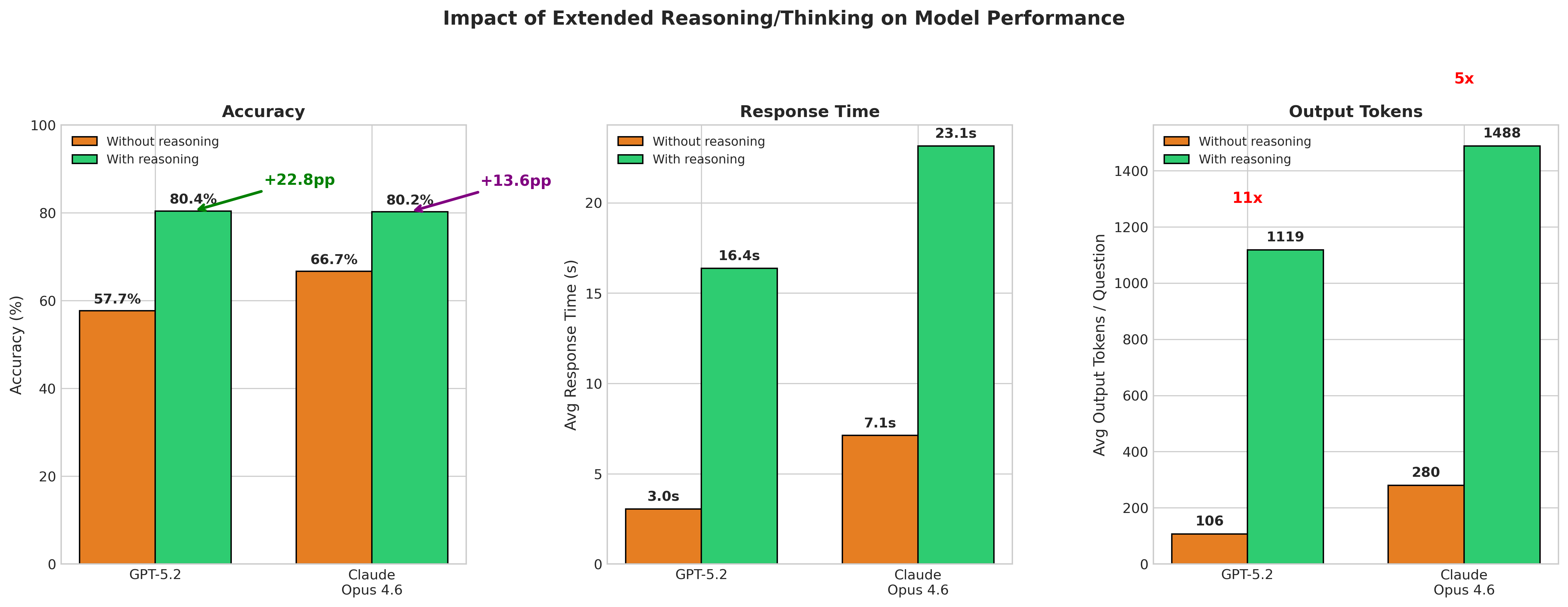}
\caption{Impact of extended reasoning/thinking on model performance across three metrics. \textbf{Left:} Accuracy comparison, where GPT-5.2 gains +22.8pp and Claude Opus 4.6 gains +13.6pp when reasoning/thinking is enabled. \textbf{Center:} Response time increases with reasoning enabled. \textbf{Right:} Output token usage per question increases substantially with reasoning (GPT-5.2: $\sim$11$\times$, Claude: $\sim$5$\times$).}
\label{fig:reasoning_impact}
\end{figure}

Figure~\ref{fig:reasoning_impact} summarizes the impact of reasoning across accuracy, response time, and token usage.

\subsection{Model Generation Progress}
An important question for practitioners is how rapidly LLM capabilities are improving on financial spreadsheet tasks. The answer, shown in Table~\ref{tab:generation_progress}, is: quickly, but not yet quickly enough. Approximately two years ago, when GPT-3.5-Turbo represented the state of the art, structured financial extraction from spreadsheets was virtually impossible; today, the best models approach 82\% accuracy, a 3.3$\times$ improvement across just two model generations. Yet despite this progress and intense cross-provider competition, with the top four models clustered within $\sim$2pp of each other, the error rates remain too high for unsupervised use. Even Gemini 3.1 Pro at 82.4\% would require substantial human review, and performance degrades further on larger, more complex files, showing that reasoning capabilities alone are insufficient to overcome the challenges of complex spreadsheet structures. The rate of improvement is fast, but standalone models are not yet production-ready.

\begin{table}[htbp]
\centering
\caption{Accuracy progression across model generations. ``vs Previous'' compares against the preceding row within the same provider, except for GPT-5.2 (reasoning), which compares against its non-reasoning variant. See Table~\ref{tab:overall_results} for GPT-3.5-Turbo's coverage breakdown.}
\label{tab:generation_progress}
\begin{tabular}{llccc}
\toprule
\textbf{Provider} & \textbf{Model} & \textbf{Coverage} & \textbf{Effective Acc.} & \textbf{vs Previous} \\
\midrule
OpenAI & GPT-3.5-Turbo$^*$ & 73.7\% & 24.8\% & --- \\
OpenAI & GPT-4o & 100\% & 54.1\% & +29.3pp \\
OpenAI & GPT-5.2 (no reasoning) & 100\% & 57.7\% & +3.6pp \\
OpenAI & GPT-5.2 (reasoning) & 100\% & \textbf{80.4\%} & +22.8pp \\
\midrule
Google & Gemini 2.0 Flash & 100\% & 50.1\% & --- \\
Google & Gemini 2.5 Pro & 100\% & 78.8\% & +28.7pp \\
Google & Gemini 3 Pro & 100\% & 80.2\% & +1.4pp \\
Google & Gemini 3.1 Pro & 100\% & \textbf{82.4\%} & +2.2pp \\
\midrule
Anthropic & Claude Opus 4.6 & 100\% & 66.7\% & --- \\
Anthropic & Claude Opus 4.6 (thinking) & 100\% & \textbf{80.2\%} & +13.6pp \\
\bottomrule
\multicolumn{5}{l}{\small $^*$Coverage = 73.7\%; accuracy on processable questions = 33.6\% (124/369).}
\end{tabular}
\end{table}

\subsection{Token Usage and Cost Analysis}

\begin{figure}[htbp]
\centering
\includegraphics[width=0.95\textwidth]{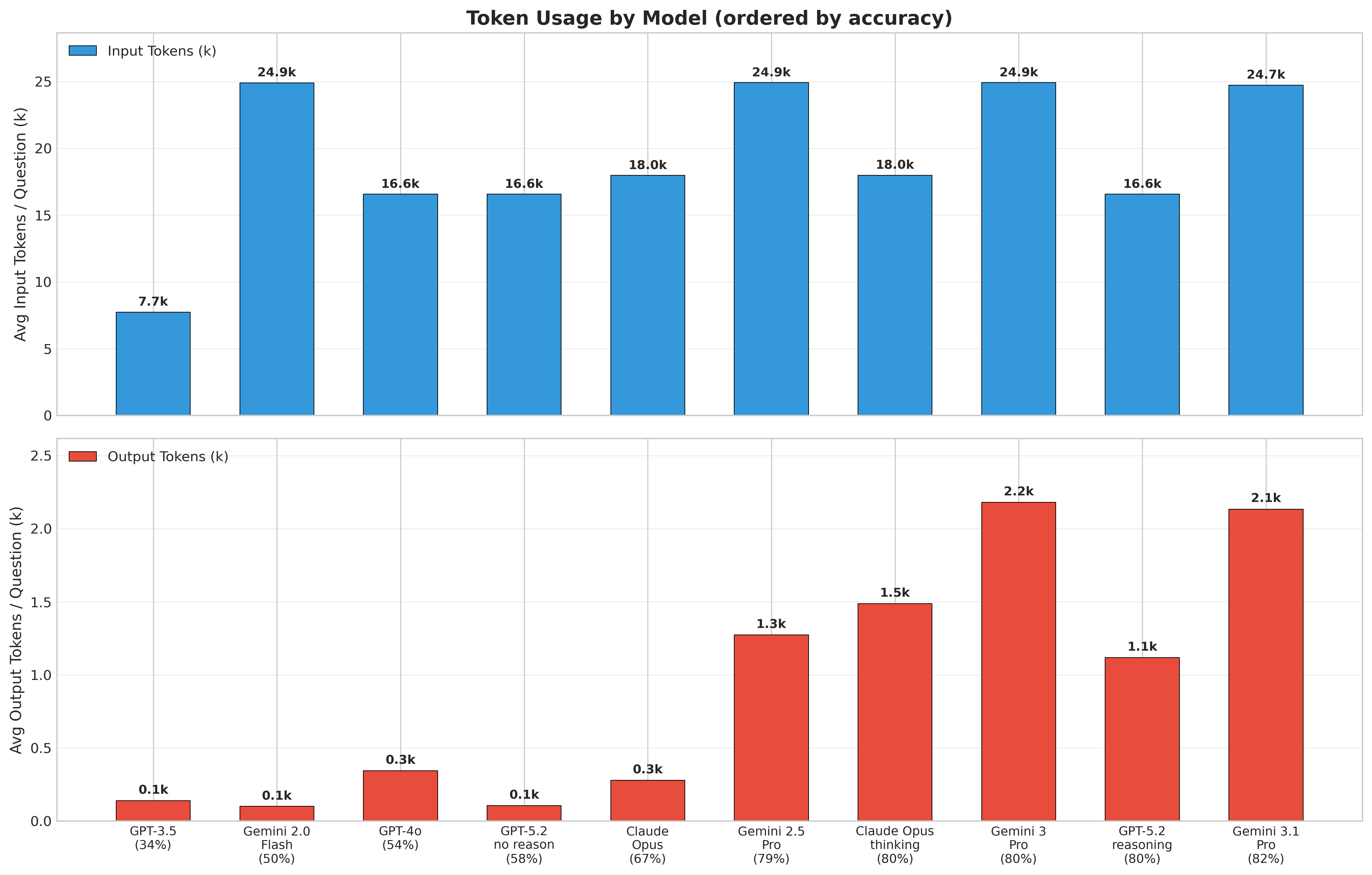}
\caption{Average input and output token usage per question by model, ordered by accuracy. Input tokens (top) and output tokens (bottom) are shown on separate scales so that the much smaller output token counts remain visible. Reasoning-enabled models consume significantly more output tokens. Claude Opus 4.6 with thinking uses $\sim$5$\times$ more output tokens than default mode.}
\label{fig:token_usage}
\end{figure}

Figure~\ref{fig:token_usage} shows the average token consumption per question for each model, and Table~\ref{tab:token_consumption} provides precise per-question statistics for cost estimation and API budget planning. Input tokens are not comparable across providers due to tokenizer differences. Thus, cross-provider comparisons are not meaningful. However, reasoning increases output tokens substantially: GPT-5.2 produces $\sim$11$\times$ more output tokens with reasoning (1,118 vs.\ 106), while Claude shows a $\sim$5$\times$ increase. This higher output volume correlates with higher accuracy in reasoning mode.

\begin{table}[htbp]
\centering
\caption{Average token consumption per question by model}
\label{tab:token_consumption}
\begin{tabular}{lrrrr}
\toprule
\textbf{Model} & \textbf{Input} & \textbf{Output} & \textbf{Total} & \textbf{Accuracy} \\
\midrule
Gemini 2.0 Flash & 24,916 & 101 & 25,017 & 50.1\% \\
Gemini 2.5 Pro & 24,917 & 1,274 & 26,191 & 78.8\% \\
Gemini 3 Pro & 24,917 & 2,180 & 27,097 & 80.2\% \\
Gemini 3.1 Pro & 24,722 & 2,133 & 26,855 & 82.4\% \\
GPT-3.5-Turbo$^*$ & 8,737 & 158 & 8,895 & 24.8\% \\
GPT-5.2 (no reasoning) & 16,567 & 106 & 16,673 & 57.7\% \\
GPT-5.2 (reasoning) & 16,567 & 1,118 & 17,685 & 80.4\% \\
GPT-4o & 16,568 & 345 & 16,913 & 54.1\% \\
Claude Opus 4.6 & 17,974 & 280 & 18,254 & 66.7\% \\
Claude Opus 4.6 (thinking) & 17,991 & 1,488 & 19,479 & 80.2\% \\
\bottomrule
\multicolumn{5}{l}{\small $^*$GPT-3.5-Turbo averages are lower because some files exceeded its 16K context limit;} \\
\multicolumn{5}{l}{\small \phantom{$^*$}token counts reflect only the processable questions.}
\end{tabular}
\end{table}

\subsection{Performance by Data File}

\begin{table}[htbp]
\centering
\caption{Accuracy by data file (all models, averaged). Version~B and~C files are structural variants with approximately one-third fewer rows and modified column layouts. For files where GPT-3.5-Turbo's evaluation returned no verification result (6 of 24 files), that model is excluded from the average.}
\label{tab:file_results}
\begin{tabular}{lccc}
\toprule
\textbf{File} & \textbf{Companies / Funds} & \textbf{Accuracy} & \textbf{Questions} \\
\midrule
synthetic1\_A.csv & 45 / 4 & 74.1\% & 22 \\
synthetic1\_B.txt & $\sim$30 / 4 & 76.8\% & 22 \\
synthetic1\_C.txt & $\sim$30 / 4 & 83.6\% & 22 \\
synthetic2\_A.csv & 89 / 5 & 69.5\% & 21 \\
synthetic2\_B.txt & $\sim$59 / 5 & 55.2\% & 21 \\
synthetic2\_C.txt & $\sim$59 / 5 & 86.2\% & 21 \\
synthetic3\_A.txt & 114 / 6 & 70.7\% & 22 \\
synthetic3\_B.txt & $\sim$76 / 6 & 79.0\% & 10 \\
synthetic3\_C.txt & $\sim$76 / 6 & 81.8\% & 22 \\
synthetic4\_A.csv & 152 / 8 & 48.6\% & 22 \\
synthetic4\_B.txt & $\sim$101 / 8 & 67.2\% & 22 \\
synthetic4\_C.txt & $\sim$101 / 8 & 63.1\% & 22 \\
synthetic5\_A.txt & 58 / 4 & 66.7\% & 21 \\
synthetic5\_B.txt & $\sim$38 / 4 & 63.3\% & 21 \\
synthetic5\_C.txt & $\sim$38 / 4 & 71.4\% & 21 \\
synthetic6\_A.txt & 46 / 5 & 75.3\% & 19 \\
synthetic6\_B.txt & $\sim$30 / 5 & 73.7\% & 19 \\
synthetic6\_C.txt & $\sim$30 / 5 & 74.2\% & 19 \\
synthetic7\_A.txt & 34 / 3 & 56.8\% & 22 \\
synthetic7\_B.txt & $\sim$22 / 3 & 62.3\% & 22 \\
synthetic7\_C.txt & $\sim$22 / 3 & 66.8\% & 22 \\
synthetic8\_A.txt & 108 / 9 & 52.0\% & 22 \\
synthetic8\_B.txt & $\sim$72 / 9 & 56.1\% & 22 \\
synthetic8\_C.txt & $\sim$72 / 9 & 49.5\% & 22 \\
\bottomrule
\end{tabular}
\end{table}

\begin{figure}[htbp]
\centering
\includegraphics[width=0.98\textwidth]{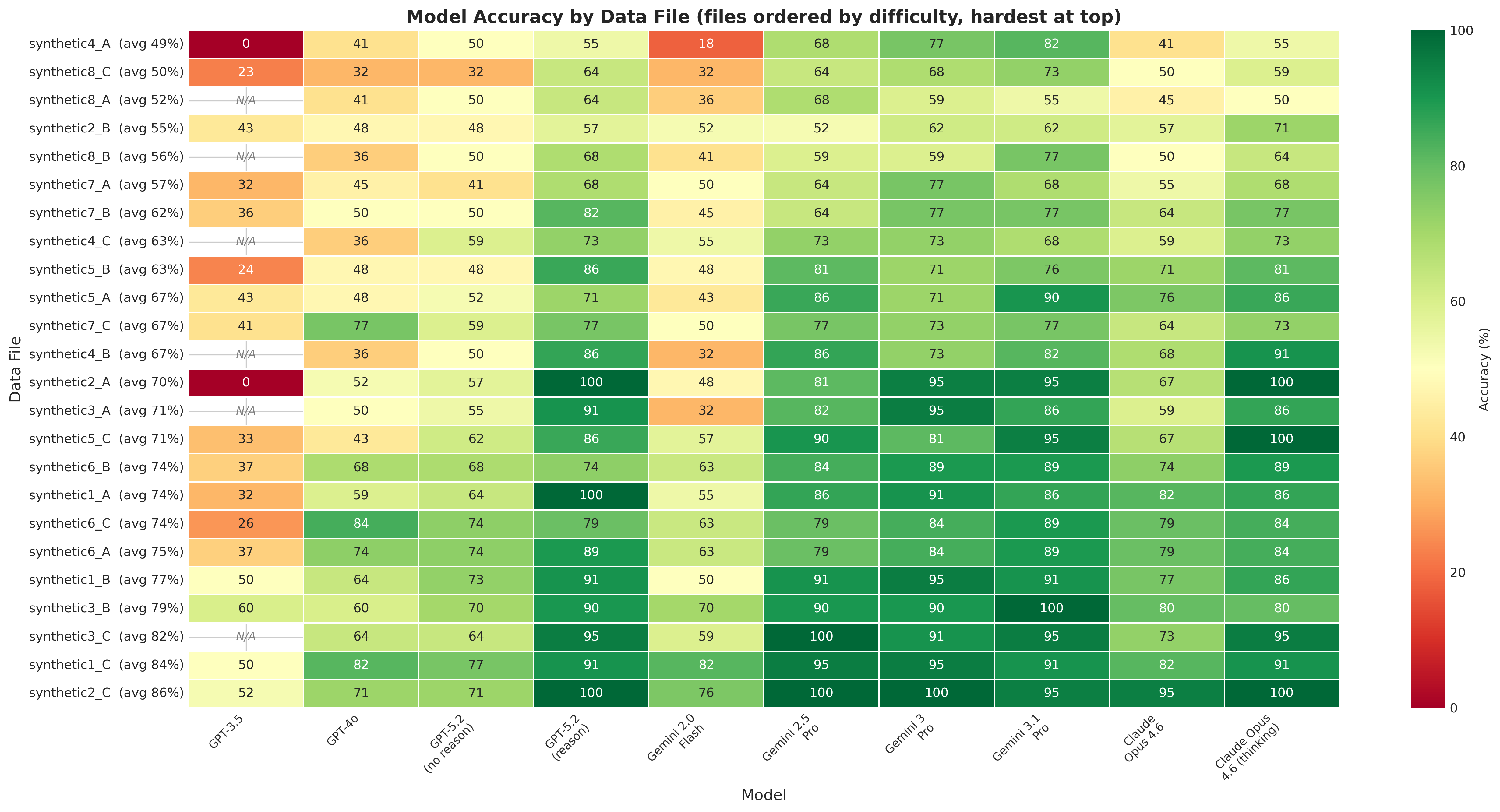}
\caption{Heatmap of accuracy by model and data file. Each cell shows the percentage of questions answered correctly by a given model on a given file. Files are ordered by average difficulty (hardest at top). The number of questions per file ranges from 10 (synthetic3\_B) to 22 (most files), with three parameterized templates sampled at \texttt{sample\_size=3}. synthetic4\_A is consistently the hardest file across all models; synthetic2\_C is among the easiest. GPT-3.5-Turbo shows no result (context overflow) on 6 of 24 files exceeding its context limit, with failures distributed across A, B, and C variants.}
\label{fig:file_heatmap}
\end{figure}

Performance varies substantially across files (Table~\ref{tab:file_results}): synthetic2\_C achieves the highest accuracy at 86.2\% while synthetic4\_A scores just 48.6\%. The drop on synthetic4\_A (152 companies, 8 funds) confirms that file complexity is a key driver of accuracy. Version~B and~C files do not uniformly outperform or underperform their A counterparts: some variants score higher (e.g., synthetic4\_B at 67.2\% vs.\ synthetic4\_A at 48.6\%, likely because the reduced row count lowers complexity; synthetic2\_C at 86.2\% vs.\ synthetic2\_A at 69.5\%) while others score lower (e.g., synthetic2\_B at 55.2\% vs.\ synthetic2\_A at 69.5\%, likely due to structural modifications like removing the fund column). Both file size and layout complexity independently affect model performance. Figure~\ref{fig:file_heatmap} visualizes these per-file differences as a heatmap across all models.

\begin{figure}[htbp]
\centering
\includegraphics[width=0.95\textwidth]{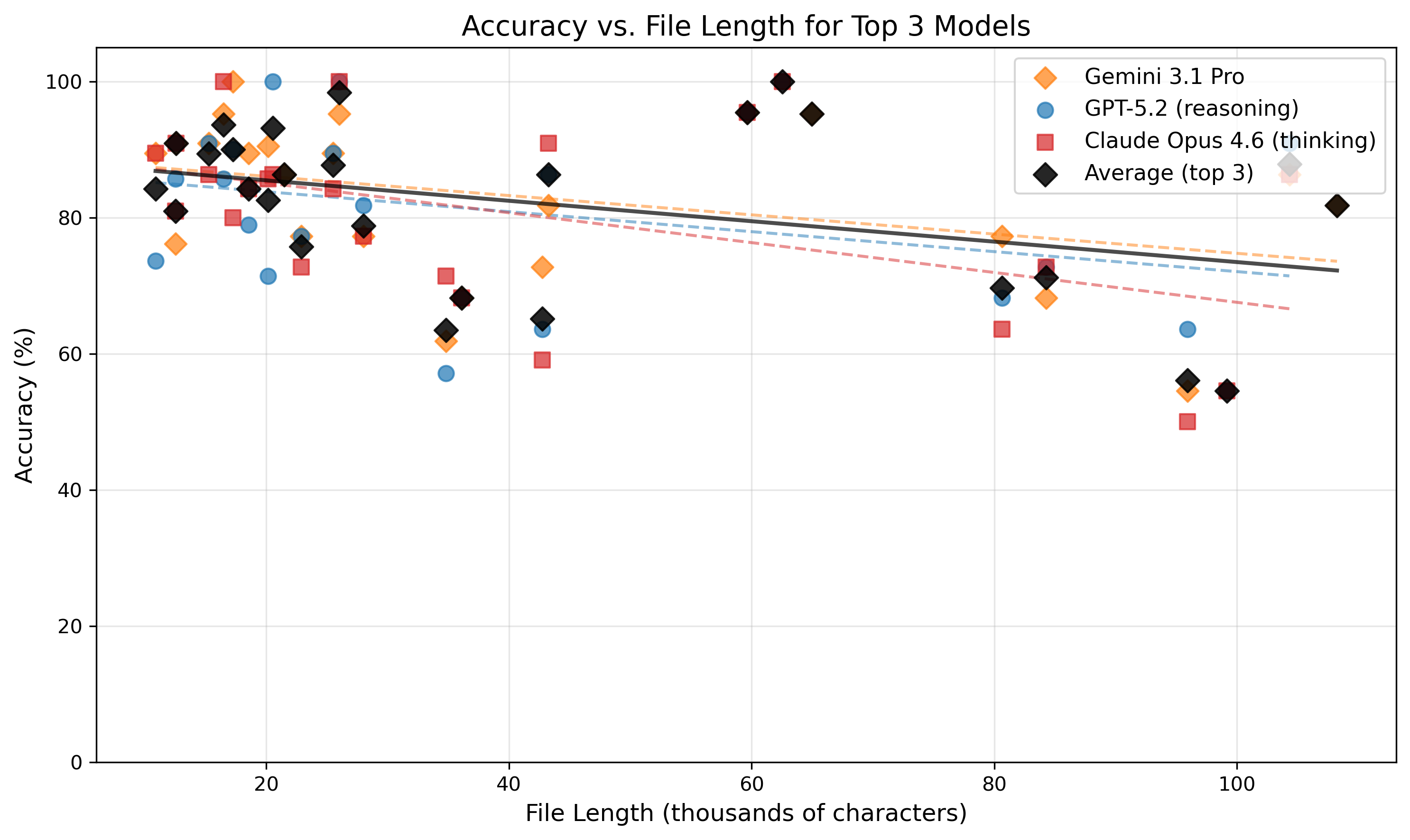}
\caption{Average accuracy of the top three models (Gemini 3.1 Pro, GPT-5.2 reasoning, Claude Opus 4.6 thinking) vs.\ file length in characters. Each point represents one evaluation file; the solid black line shows the average trend. The negative slope ($-0.23$~pp per 1K characters, $R^{2} = 0.30$) indicates a moderate relationship, though file-specific structural complexity remains an important confounding factor.}
\label{fig:accuracy_vs_length}
\end{figure}

\paragraph{File length vs.\ accuracy.}
Figure~\ref{fig:accuracy_vs_length} examines whether raw file length (in characters) predicts accuracy for the top three models (Gemini 3.1 Pro, GPT-5.2 with reasoning, Claude Opus 4.6 with thinking). The linear trend is negative but moderate: approximately $-0.23$~percentage points per 1K additional characters ($R^{2} = 0.30$, Pearson $r = -0.55$). While accuracy does tend to be lower on the longest files (e.g., synthetic4\_A at 108K characters averages 64\% across the top three models, vs.\ 91\% on synthetic1\_C at 13K characters), the relationship is confounded by the fact that longer files also tend to have more funds, more complex layouts, and different structural modifications. Shorter files that happen to have unusual structural changes (e.g., removing the fund column) can perform worse than longer but structurally simpler files.

\section{Discussion}

\subsection{Accuracy Requirements in Financial Due Diligence}

None of the ten configurations tested produce error rates acceptable for unsupervised production use. Accuracy drops from above $\sim$90\% on simple lookups to just above 30\% on complex calculations for the top 3 models, and this gradient is shared across all models, pointing to a structural limitation of LLM-based numerical reasoning on tabular data.

This limitation matters in alternative investment due diligence, where extracted data feeds directly into investment decisions. An LP evaluating a \$50M commitment to a private equity fund relies on accurate portfolio metrics (IRR, multiples, enterprise values) to assess fund performance. A 5\% error rate in data extraction could lead to systematic misvaluation, erroneous peer comparisons, and poor capital allocation decisions.

\subsection{Why Standalone LLMs Fail on Spreadsheets}

Our results point to several likely limitations:

\begin{itemize}
    \item \textbf{Tokenization disrupts structure}: Sequential tokenization disrupts the two-dimensional structure of spreadsheets. Column headers become separated from their values, and row relationships are difficult to maintain across long contexts.
    \item \textbf{Position vs.\ semantics}: Spreadsheets encode meaning through position (cell C5 derives meaning from column header C1 and row label A5), a spatial relationship that token-based processing handles poorly. Our text serialization (Section~\ref{sec:serialization}) removes much of this structure before the model sees it; alternative modalities (e.g., rendered images) could mitigate this.
    \item \textbf{Numeric precision}: LLMs lack specialized numerical processing and cannot match deterministic computation, particularly for financial ratios with multiple decimal places.
    \item \textbf{Modality-dependent vs.\ modality-independent failures}: It is useful to distinguish failures that are \emph{caused by} text serialization (e.g., row misidentification due to lost visual cues) from failures that are \emph{inherent to LLM reasoning} regardless of input modality (e.g., incorrect arithmetic, sorting errors, median computation). Our category-level results (Table~\ref{tab:question_types}) suggest the latter dominate: sorting, complex aggregation, and counting fail at high rates.
\end{itemize}

\subsection{Hypothetical: Brute-Force Extraction via Per-Value Queries}

Our benchmark results suggest an interesting hypothetical approach to closing the accuracy gap: since the best-performing models achieve near-perfect accuracy on single-value lookups (e.g., Q10 entry enterprise value: 100\% for four of the top five models), one could in principle reconstruct an entire standardized table by issuing one LLM call per cell, asking for each company's value in each column individually, and then performing all calculations deterministically on the resulting table. This ``brute-force extraction'' approach would bypass the multi-step reasoning failures that dominate our error analysis (sorting, aggregation, median computation) by reducing every extraction to the task category where LLMs already excel: simple lookups.

However, this approach has significant practical limitations. First, it requires a \emph{structural discovery} step before per-value extraction can begin: the system must know which companies exist, which funds they belong to, and which columns are present, itself a non-trivial task (Q1 ``how many funds?'' reaches only 83\% even for top models, and Q2 ``how many companies per fund?'' achieves 74\% for GPT-5.2 with reasoning). Second, the cost would be extreme: a spreadsheet with 100 companies and 15 data columns would require $\sim$1{,}500 individual API calls, each sending the full spreadsheet context ($\sim$15--20K input tokens), resulting in $\sim$30M input tokens for a single file. At current API pricing, this would cost roughly \$100 per spreadsheet depending on the provider and model, compared to a few cents for a single-call approach that answers all questions at once. Third, latency would be prohibitive: at $\sim$5 seconds per call, 1{,}500 calls would take over 2 hours sequentially; even with parallelization, rate limits would make this impractical for production use on hundreds of files.

Nevertheless, the brute-force approach is conceptually instructive: it demonstrates that \textbf{decomposing extraction from computation} is the core architectural insight. If each value can be retrieved individually with near-perfect accuracy, and all downstream calculations are performed deterministically, the accuracy degradation on complex tasks disappears entirely.

A more practical middle ground between single-call QA and per-cell brute force is a \textbf{staged pipeline} approach:

\begin{enumerate}
    \item \textbf{Schema discovery}: Use one or two LLM calls to identify the table headers, fund boundaries, and column semantics, producing a structured schema (e.g., a JSON column map).
    \item \textbf{Row-level extraction}: For each row (company), issue a single LLM call that extracts all column values at once, guided by the discovered schema. This reduces the per-file call count from $\sim$1{,}500 (one per cell) to $\sim$100 (one per company) for a 100-company file.
    \item \textbf{Deterministic computation}: All downstream calculations (averages, medians, ratios, sorting) are performed programmatically on the extracted table, eliminating the multi-step reasoning failures that dominate our error analysis.
\end{enumerate}

\noindent This staged approach exploits the near-perfect single-value lookup accuracy we observe while avoiding both the cost explosion of per-cell queries and the reasoning failures of single-call QA on complex questions. It also produces an auditable intermediate artifact (the extracted table) that supports human review and regulatory transparency. Evaluating such pipeline architectures is a key direction for future work. Given the error rates observed, human review remains essential for high-stakes decisions. Automated extraction can reduce mechanical effort, freeing analysts for interpretation, but cannot replace verification.

\section{Limitations}
\label{sec:limitations}

Our study has several limitations:

\begin{itemize}
    \item All results reflect LLM performance on \emph{text-serialized} spreadsheets (comma-separated text), not on native Excel files. The serialization discards merged cells, cross-sheet references, and visual formatting cues (bold text, cell colors, table borders) that carry semantic meaning in financial spreadsheets (see Section~\ref{sec:serialization}). While text serialization is the industry-standard approach, results with alternative input modalities (rendered images, structured JSON, tabular HTML) could differ materially. Vision-capable models could potentially recover spatial cues, though they face their own challenges with dense numeric tables at scale.
    \item The benchmark uses synthetic data modeled on real structural templates, which may not capture all real-world complexity (e.g., embedded charts or data split in multiple sheets).
    \item The benchmark focuses on private equity; other financial domains may present different challenges.
    \item With \texttt{sample\_size=3} for parameterized templates, per-file and per-category sample sizes are small.
\end{itemize}

\section{Conclusion}

We have introduced FinSheet-Bench, a benchmark for evaluating LLM performance on financial spreadsheet comprehension, and through evaluation of ten model configurations we demonstrate that no standalone LLM achieves error rates low enough for unsupervised use in professional finance. Progress has been rapid: tasks that seemed entirely out of reach two years ago, when GPT-3.5-Turbo was the state of the art and achieved just 33.6\% accuracy on processable questions, are now at the edge of feasibility, with the best models approaching 82\%. However, the consistent accuracy degradation we observe across all models, from $\sim$89\% on simple lookups to $\sim$20\% on complex aggregations, shows a clear limitation: LLMs handle content retrieval well but struggle with multi-step numerical reasoning on tabular data. Reliable financial spreadsheet extraction will likely require architectural approaches that separate document understanding from deterministic computation, rather than relying on ever-more-powerful standalone models. For the alternative investment industry, where due diligence requires processing hundreds of non-standardized spreadsheets per year, the benchmark provides a standardized evaluation framework to measure progress toward production-ready extraction.

\subsection{Future Work}

We plan to extend this work in several directions:

\begin{itemize}
    \item \textbf{Expanded benchmark}: Adding additional structural variations (merged cells, multiple sheets, embedded charts) to further stress-test extraction capabilities. All eight A-version, eight B-version, and eight C-version files have been benchmarked.
    \item \textbf{Format sensitivity and multimodal input}: Converting FinSheet-Bench files to alternative representations (structured JSON, tabular HTML, rendered images) to analyze how data presentation affects accuracy across providers. In particular, evaluating vision-capable models on rendered spreadsheet images could recover visual formatting cues (bold text, underlines, cell borders, background colors) that text serialization destroys, potentially reducing parsing errors, though we expect the core numerical reasoning failures (sorting, complex aggregation) to persist regardless of input modality.
    \item \textbf{Parsing vs.\ calculation decomposition}: Controlled experiments to disentangle parsing failures (wrong column/row selection due to serialization) from calculation failures (incorrect arithmetic on correctly identified values), enabling more targeted improvements to both serialization methods and model capabilities.
    \item \textbf{Agentic and pipeline-based approaches}: Evaluating agentic architectures (tool-augmented agents, code execution approaches, multi-stage extraction pipelines) that decompose document understanding from numerical computation, motivated by the gap our results expose between content retrieval (where LLMs already work well) and multi-step calculation (where they do not).
\end{itemize}

\section*{Author Contributions}

\textbf{Jan Ravnik} designed and ran the benchmarking experiments, developed the benchmarking framework, and wrote the paper.
\textbf{Matja\v{z} Li\v{c}en} prepared the synthetic portfolio datasets used in FinSheet-Bench and assisted in writing the paper.
\textbf{Felix B\"uhrmann} contributed to experiment design.
\textbf{Bithiah Yuan} contributed to the research design and methodology.
\textbf{Felix Stinson} contributed financial domain expertise, helped design financially relevant benchmark questions, and confirmed the validity of the synthetic financial data.
\textbf{Tanvi Singh} supervised the project, contributed to experiment design and ideation, and assisted in writing the paper.

\FloatBarrier
\section*{Acknowledgments}

We thank the anonymous reviewers for their feedback. This work was partially supported by an Innosuisse Innovation Cheque (\url{https://www.innosuisse.admin.ch/en/innovation-cheque}).

Jan Ravnik acknowledges the use of AI-based writing aids to assist with language and phrasing during the preparation of this manuscript.

\bibliographystyle{plainnat}

\appendix

\section{Version B and C Modifications}
\label{app:version_details}

Each of the 8 base files (version~A) was used to derive two structural variants (versions~B and~C). While the underlying data remained mostly intact, approximately one-third of rows were removed at random and various structural modifications were applied. The specific changes for each file are listed below.

\subsection{Synthetic~1}

\textbf{Version B:}
\begin{itemize}
    \item Split city and state into two columns
    \item Fund listed in a separate column
    \item Added separate columns for Exit and Current (split from Exit/Current)
    \item Removed empty columns between categories
\end{itemize}

\textbf{Version C:}
\begin{itemize}
    \item Added average and median rows below each fund
    \item Added Net Debt/EBITDA columns
    \item Merged various header cells
    \item Deleted Net Multiple and IRR columns
\end{itemize}

\subsection{Synthetic~2}

\textbf{Version B:}
\begin{itemize}
    \item Removed Fund column and moved fund information to separator rows
    \item Removed Exit and Current columns (kept only Exit/Current columns)
    \item Deal Team Members separated by commas instead of semicolons
    \item Removed average rows at the bottom
\end{itemize}

\textbf{Version C:}
\begin{itemize}
    \item Removed Fund column and moved fund information to separator rows
    \item Added average values under each fund
    \item Added blank columns between column categories
\end{itemize}

\subsection{Synthetic~3}

\textbf{Version B:}
\begin{itemize}
    \item Merged and centered header across multiple rows
    \item Removed a significant number of columns; currency now denoted as DKK at the top of the sheet
    \item Exit / 2024 / 2025 statistics now denoted in the header
    \item Fund denoted as watermark text behind respective cells
    \item Added blank rows between funds
    \item Deleted Gross MOIC column
\end{itemize}

\textbf{Version C:}
\begin{itemize}
    \item Moved fund information from column to separator rows
    \item Fund rows now contain averages and other summary statistics
    \item Replaced semicolons with commas as list separators
    \item Merged header across multiple rows
\end{itemize}

\subsection{Synthetic~4}

\textbf{Version B:}
\begin{itemize}
    \item Dropped a significant portion of columns
    \item Moved all entry and exit values so that they are grouped together
    \item Added blank rows between funds (but kept fund column)
    \item Added blank columns between column categories
\end{itemize}

\textbf{Version C:}
\begin{itemize}
    \item Removed Fund column and moved fund information to separator rows
    \item Added Average, Median, Max, and Min summary rows for every fund
    \item Exit date: replaced blank values with current date
    \item Split city and state into two columns
\end{itemize}

\subsection{Synthetic~5}

\textbf{Version B:}
\begin{itemize}
    \item Board representatives separated by commas instead of semicolons
    \item Funds denoted in a vertically merged cell (spanning multiple rows) and separated by different cell fill colors
    \item ``Exit date'' moved under Exit category and renamed to ``Date''
    \item ``Entry date'' moved under Entry category and renamed to ``Date''
\end{itemize}

\textbf{Version C:}
\begin{itemize}
    \item Data ordered by sector with blank rows between sectors
    \item Deselected the ``Filter'' option (Excel feature only)
    \item Added ``Current/At Exit'' category
    \item Added blank columns between categories
\end{itemize}

\subsection{Synthetic~6}

\textbf{Version B:}
\begin{itemize}
    \item Removed average rows under each fund
    \item Added an intentional typo: ``Fund II'' $\rightarrow$ ``Fdund II'' (the label appears only once in the sheet)
    \item Merged some columns in the header
\end{itemize}

\textbf{Version C:}
\begin{itemize}
    \item Replaced enumerated fund names with descriptive words: Fund~I $\rightarrow$ Equity, Fund~II $\rightarrow$ Diversify, Fund~III $\rightarrow$ Income, Fund~IV $\rightarrow$ Growth, Fund~V $\rightarrow$ Stability
    \item Added blank columns between categories
\end{itemize}

\subsection{Synthetic~7}

\textbf{Version B:}
\begin{itemize}
    \item Replaced enumerated fund names with descriptive words: Fund~I $\rightarrow$ Income, Fund~II $\rightarrow$ Growth, Fund~III $\rightarrow$ Stability
    \item Added blank rows, removed blank columns
\end{itemize}

\textbf{Version C:}
\begin{itemize}
    \item Randomly shuffled rows within each fund so they are no longer ordered by date
    \item Removed Fund column and replaced it with fund annotation rows
\end{itemize}

\subsection{Synthetic~8}

\textbf{Version B:}
\begin{itemize}
    \item Replaced semicolons with commas as list separators
    \item Added ``City'', ``State'', ``Country'', and ``Currency'' columns
    \item Added ``Average'', ``Median'', ``Max'', and ``Min'' summary rows at the bottom of all entries
    \item Added blank rows between funds (skipped for Fund~I and Fund~II as they have mixed entries labelled ``Fund I \& II'')
\end{itemize}

\textbf{Version C:}
\begin{itemize}
    \item Shuffled column order
    \item Merged and centered headers
    \item Exit and entry dates split into three columns: year, month, day
    \item Removed notes column and notes
\end{itemize}

\end{document}